\documentclass{article}

\usepackage{arxiv}

\usepackage[utf8]{inputenc} 
\usepackage[T1]{fontenc}    
\usepackage{hyperref}       
\usepackage{url}            
\usepackage{booktabs}       
\usepackage{amsfonts}       
\usepackage{nicefrac}       
\usepackage{microtype}      
\usepackage{lipsum}		
\usepackage{graphicx}
\usepackage{natbib}
\usepackage{doi}

\usepackage{mathtools,amsmath,amssymb}
\usepackage{algorithmicx, algorithm}
\usepackage[noend]{algpseudocode} 

\usepackage{amsthm, amssymb}
\theoremstyle{definition}
\newtheorem{definition}{Definition}[section]

\theoremstyle{proposition}
\newtheorem{proposition}{Proposition}[section]

\title{A Graph Neural Network-Based QUBO-Formulated Hamiltonian-Inspired Loss Function for Combinatorial Optimization using Reinforcement Learning}

\author{ \href{https://orcid.org/0000-0000-0000-0000}{\includegraphics[scale=0.06]{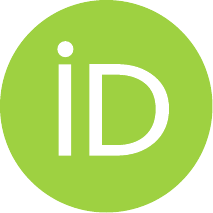}\hspace{1mm}Redwan Ahmed Rizvee} \\
	Department of Computer Science and Engineering\\
	University of Dhaka\\
	Dhaka, Bangladesh \\
	\texttt{rizvee@du.ac.bd} \\
	\And
	\href{https://orcid.org/0000-0000-0000-0000}{\includegraphics[scale=0.06]{orcid.pdf}\hspace{1mm}Md. Mosaddek Khan} \\
	Department of Computer Science and Engineering\\
	University of Dhaka\\
	Dhaka, Bangladesh\\
	\texttt{mosaddek@du.ac.bd}\\
}

\renewcommand{\shorttitle}{\textit{arXiv} Template}

\hypersetup{
pdftitle={Extension to: Combinatorial optimization with physics-inspired graph neural networks through Reinforcement Learning Based setups},
pdfsubject={q-bio.NC, q-bio.QM},
pdfauthor={David S.~Hippocampus, Elias D.~Striatum},
pdfkeywords={First keyword, Second keyword, More},
}

\begin{document}
\maketitle

\begin{abstract}
Quadratic Unconstrained Binary Optimization (QUBO) is a generic technique to model various NP-hard combinatorial optimization problems in the form of binary variables. The Hamiltonian function is often used to formulate QUBO problems where it is used as the objective function in the context of optimization. Recently, PI-GNN, a generic scalable framework, has been proposed to address the Combinatorial Optimization (CO) problems over graphs based on a simple Graph Neural Network (GNN) architecture. Their novel contribution was a generic QUBO-formulated Hamiltonian-inspired loss function that was optimized using GNN. In this study, we address a crucial issue related to the aforementioned setup especially observed in denser graphs. The reinforcement learning-based paradigm has also been widely used to address numerous CO problems. Here we also formulate and empirically evaluate the compatibility of the QUBO-formulated Hamiltonian as the generic reward function in the Reinforcement Learning paradigm to directly integrate the actual node projection status during training as the form of rewards. In our experiments, we observed up to $44\%$ improvement in the RL-based setup compared to the PI-GNN algorithm. Our implementation can be found in \footnote{\url{https://github.com/rizveeredwan/learning-graph-structure/}}.
\end{abstract}


\section{Introduction and Motivation}
Combinatorial Optimization (CO) is a branch of optimization that deals with finding the best solution from a finite pool of possibilities where the chosen solution maintains the given set of constraints and optimizes a problem-specific objective. However, due to the size of the problem's variables and the corresponding large size of the set of possible solutions, it is difficult to find an exact solution which leads to finding an approximate one. \cite{schuetz2022combinatorial} proposed a Graph Neural Network (GNN) based solution (PI-GNN) to address the Combinatorial Optimization (CO) problems over graphs. First, they formulate the problem using a Quadratic Unconstrained Binary Optimization (QUBO). Then based on that they apply a generic loss function over stacked layers of GNN to produce the node probability distribution which leads to the labeling of all the nodes as $\{0, 1\}$. This labeling as either $0$ or $1$ denotes the set of nodes (or edges) that will belong to the solution to optimize the problem's objective or reduce the number of violated constraints. The main conceptual novelty of PI-GNN is generality and scalability.

Recently, \cite{boettcher2023inability} demonstrated in a critical review paper that PI-GNN performs worse than traditional greedy algorithms when solving the Max-Cut problem. Additionally, \cite{angelini2023modern} raises a similar kind of concern that the addressed simple GNN-based solution lacks behind classical greedy algorithms presenting their discussion over the Maximum Independent Set problem. Both of the literature mainly expressed that, in a very specifically curated problem (e.g., Max-Cut, Maximum Independent Set, etc.) the relevant greedy algorithms might work better than PI-GNN. Though the author of the PI-GNN has argued over the comments \citep{boettcher2023inability, schuetz2023reply} that ignoring the generality and scalability of their proposed framework reduces the merit of their work. They also provided empirical results to support the argument. In this study, we also support the generality and scalability of PI-GNN, especially their loss function formulation based on QUBO-formulated Hamiltonian. However, we also highlight a concern that we investigate in this work. We highlight the concerns through the following points,

\begin{enumerate}
    \item Absence of actual projection Strategy in loss function: The actual projection or node labeling strategy is not present in the generic loss function stated in \cite{schuetz2022combinatorial}. Apart from the interpretability of the node labeling, this also raises an important concern. Using gradient descent optimization the architecture might converge to a local minima but while projecting the actual node labels the performance may significantly deteriorate due to the absence of an actual node projection strategy in the loss function. The projected node labels represent the actual constraint satisfaction status of the graph in the current iteration which should guide the loss function as per our observation.
    \item Implementational issue in denser graphs: While experimenting with PI-GNN in graphs of different densities, we observed a crucial concern \footnote{\url{https://github.com/amazon-science/co-with-gnns-example}}, mostly in denser graphs. In the implementation to early stop the training, they track consecutive loss variation with a patience value $p (p=100)$ and tolerance value $\tau (=10^{-4})$. When there is no consecutive loss reduction for $p$ epochs or the consecutive loss variation is less than $\tau$ occurs for $p$ epochs, the training is stopped. In our experiments, we observed some quality performance degradation in denser graphs due to this early stopping strategy.
\end{enumerate}

Based on the aforementioned observations, we summarize the following contributions of this article,

\begin{enumerate}
    \item Fuzzy Early stopping strategy: We suggest applying some relaxed or fuzzy early stopping strategies. We suggest omitting $\tau$ and tracking improvement over the best objective function value $obj^{*}$ observed till this phase of the iteration. If no improvement is observed for $p$ epochs over $obj^{*}$, we can stop the training. In denser graphs, based on the early stopping strategy used in the official implementation, we often received a uniform node probability distribution leading to a significant number of constraint violations. However, by applying the suggested relaxed strategy we generated a quality node probability distribution leading to the number of significantly reduced number of constraint violations with PI-GNN.

    \item Compatibility of QUBO-formulated Hamiltonian as generic reward function in Reinforcement Learning (RL) based setup: To embed the actual projection status in the loss function or loss objective during training we experiment with the QUBO-formulated Hamiltonian as a generic reward function in RL-based setups. Being inspired by the formulation stated in \cite{drori2020learning} and \cite{khalil2017learning}, we establish a modified generic framework $GRL$ that works with the QUBO-formulated Hamiltonian as the generic reward function. Additionally, we also formulate a Monty Carlo Tree Search with a GNN-based solution where we apply a guided search through manual perturbation of node labels during training. Empirically, we have gained up to $44\%$ improvement in reducing the number of violated constraints.
\end{enumerate}

To continue the discussion of this study, we address the Max-Cut problem. But all the presented proposals are generic in manner and can be extended to a wide group of graph-based canonical optimization problems as stated in \cite{schuetz2022combinatorial}. In simple terms, the classical Max-Cut problem targets to divide the nodes of an undirected unweighted graph into two different sets such that the number of ``cut edges" maximizes where a node is in one set and the other node of the edge is in the other set.

The rest of the study is organized as follows. In Section \ref{section:designed_methodology}, we present all our proposals and formulations that we have experimented with in our work. In Section \ref{section:evaluation}, we empirically evaluate all the architectures with PI-GNN based on various metrics and discuss the results. Finally, we conclude the study through an overall review and direct toward the future extensions in Section \ref{section:conclusion}.

\section{Our Proposals} \label{section:designed_methodology}
In this section, we present all the proposals and the formulations presented in this article in a concise manner.

\subsection{PI-GNN with Fuzzy Strategy}
In this section, first, we formally present two early stopping strategies that can be used during the training of PI-GNN \cite{schuetz2022combinatorial}. Then, we discuss the main distinguishable differences between them and the underlying reasonings. For the sake of discussion, we name the strategies as, \textit{Strict Stopping} and \textit{Fuzzy Strategy}.

\begin{definition}[Strict Stopping] \label{def:strict_stop}
    During training, for the objective function $F_{obj}$, if no successive improvement is observed for consecutive $p$ epochs or the successive variation is lesser than $\tau$ occurs for consecutive $p$ epochs, the training can be stopped.
\end{definition}

\begin{definition}[Fuzzy Stopping] \label{def:fuzzy_stop}
    Let us assume that, $obj^{*}$ denotes the current best value observed for the objective function $F_{obj}$ during any phase of the training iterations. If no improvement in the value of $F_{obj}$ occurs for successive $p$ epochs compared to $obj^{*}$, the training can be stopped.
\end{definition}

Definition \ref{def:fuzzy_stop} is much fuzzier compared to the definition \ref{def:strict_stop}. Because, during the training with gradient descent, it is a quite common phenomenon that sometimes the objective function variates quite slowly before moving into larger reductions. Definition \ref{def:fuzzy_stop} addresses this by removing the dependency over $\tau$. During the training of PI-GNN, a common phenomenon is that the objective or the loss function starts from a high positive value and then with gradual training moves into larger negative values. Especially, in denser graphs, we often observed this occurrence, during the transition from positive loss to negative loss, a period occurs when the loss varies or improves a bit very insignificantly (e.g., $ < 10^{-7}, < 10^{-8}$, etc.). Then the $\tau$ value or the early stopping can degrade the performance quite significantly.

\subsection{PI-GNN Architecture and Loss function}
PI-GNN \citep{schuetz2022combinatorial} consists of k layers of stacked GNN and the last layer is a sigmoid layer to generate the node probability distributions. PI-GNN takes the QUBO-based Hamiltonian formulation ($Q$) of a problem, e.g. $F_{obj}(X) = X^{T}Q{X}=\sum_{i \leq j} x_{i}Q_{ij}x_{j}$ where $x_{i}$ denotes the variable related to the $i^{th}$ node of the problem and $Q$ denotes the problem encoded matrix. The task of PI-GNN is to set the values of $x_{i}$ variables to either $0$ or $1$ based on training to maximize $F_{obj}(X)$.

PI-GNN develops a loss function based on the QUBO-formulated Hamiltonian and minimizes the function as stated in equation \ref{eqn:pi_gnn_loss}. Here $P_{i}(\theta)$ denotes the PI-GNN generated node probability for the $i^{th}$ node. The modification over the base article that we apply is the usage of Fuzzy stopping during termination.

\begin{equation} \label{eqn:pi_gnn_loss}
    \text{maximize} \sum_{i \leq j} x_{i} Q_{ij} x_{j}  \,\to\,  \text{minimize}(-\sum_{i \leq j} P_{i}(\theta) Q_{ij} P_{j}(\theta))
\end{equation}

\subsection{Generic Reinforcement Learning framework, $GRL$} \label{section:grl}
\cite{drori2020learning} proposes a generic RL-based framework to address a wide range of combinatorial optimization problems. They use a Graph Attention Network (GAT) based encoder architecture to encode and generate the node feature vectors, upon which they apply an attention-based decoding mechanism to greedily select the nodes and apply the node labelings.

We formulate a modified version of \cite{drori2020learning} in this study. The main differences are pointed out as follows,

\begin{enumerate}
    \item Generic QUBO-formulated Hamiltonian Reward Function: A subset of terms from $X^{T}QX=\sum_{i \leq j}x_{i} Q_{ij}x_{j}$ is considered the observed reward $r^{t}$ at time $t$ during training for a particular epoch $e$. When a node $v_{i}$ is greedily selected and labeled, we check which terms $x_{i}Q_{ij}x_{j}$ where $i \leq j$ and $x_{j}Q_{ji}x_{i}$ where $j < i$ can be calculated and sum them. This is considered as the reward, $r^{t} = \sum_{i \leq j} x_{i}Q_{ij}x_{j} + \sum_{j < i}x_{j}Q_{ji}x_{i}$.
    \item Attention-based decoding strategy: In equation \ref{eqn:decoder_attn_modified}, we provide the mathematical formulation to calculate the attention or weight for node $v_{i}$ as $\gamma_{i}$. Here, $\phi_{1} \in \mathbb{R}^{d_{h} \times d}, \phi_{2}\in \mathbb{R}^{d_{h} \times d}$ and $\phi_{3}\in \mathbb{R}^{d_{h} \times n}$ are weights or architecture parameters. $\mu_{i}$ denotes the node feature vector (row vector) for the node $v_{i}$ reported from the GAT layer. $C$, $d$ and $d_{h}$ all are hyperparameters. $n$ denotes the number of nodes in the input graph. Through applying a sigmoid non-linear activation over $\gamma_{i}$, node probability distribution $P_{i}(\theta)$ is generated. Over $P_{i}(\theta)$, a probability threshold $\beta$ ($e.g., \beta=0.5$) is applied to fix the node labels, e.g., $P_{i}(\theta) \geq \beta \text{ leads to} x_{i}=1$. Here $\theta$ denotes the complete set of trainable architecture parameters. To denote the dimensions of the weights (e.g., $\phi$) we maintain output dimension times input dimension all throughout the study. $T$ is used to denote the transpose of a vector (row to column or column to row).

    \begin{equation}
    \label{eqn:decoder_attn_modified}
    \gamma_{i} =  C \text{ tanh}(\frac{{ \mu_{i}\phi_{1}^{T} .(X_{v}\phi_{3}^{T} + \sum_{j \in N(i)} \mu_{j}\phi_{2}^{T}) }}{\sqrt{d_{h}}})
\end{equation}

     To update a node $v_{i}$'s attention weight $\gamma_{i}$ we consider three aspects, node $v_{i}$'s own node feature vector ($\mu_{i}$), its adjacent neighbors' ($N(i)$) feature vectors ($\mu_{j} \text{ where } j \in N(i)$) and the current condition of node labeling $X_{v}$. If a node $v_{j}$ has already been selected and labeled then the $j^{th}$ entry will be $1$ otherwise it will remain as $0$. So $X_{v}$ is a binary vector. This formation is comparatively more contextual than the decoding formation stated in \cite{drori2020learning}. In each selection, as the greedy choice, from the pool of unselected nodes, the node with the highest attention value $\gamma_{i}$ was selected and labeled. When a node $v_{i}$ is selected, its neighbors ($j \in N(i)$) attention weights $\gamma_{j}$ are updated.
\end{enumerate}

Now, we point out the common strategies that we adopt from \cite{drori2020learning} through the following points,

\begin{enumerate}
    \item GAT as an encoder architecture: $K$ stacked layers of GAT are used as the encoder architecture to generate the node feature vectors $\mu$. The last layer consists of a sigmoid non-linear activation layer to generate the node probabilities. This probability helps to select the first node to initiate the decoding with the attention mechanism.
    \item Loss objective and Training: The loss objective that we use is stated in equation \ref{eqn:decoder_attn_modified}. Here $P(v^{t})$ denotes the probability of the greedily selected node $v$ at the $t^{th}$ iteration. $r^{t}$ has already been defined prior. $b$ denotes the reward observed from the baseline architecture at the $t^{th}$ iteration. In our setup, we do not use any baseline architecture. After accumulating the complete set of rewards (termination of an epoch), we apply gradient descent over the model parameters and backpropagate.

    \begin{equation}
        L(\theta) = \sum_{t=1}^{n} (r^{t}-b) \times P(a^{t})
         =  \sum_{t=1}^{n} r^{t} \times P(a^{t}) [\text{when $b=0$}]
    \end{equation}
\end{enumerate}

\subsection{Monty Carlo Tree Search with GNN through manual perturbation, MCTS-GNN} \label{section:mcts_gnn}
In this section, we concisely present a formulation where we integrate the Monty Carlo Tree Search in assistance with GNN in an RL-based setup. The main idea is that each node of the search tree consists of a partial solution (or node labels), and based on that a single GNN is trained to approximate the remaining nodes' labels. The goal is, using this manual perturbation of node labels, a guided search is conducted to maximize the amount of rewards. Now, we discuss the strategies based on RL terminologies (state, action, reward) and Monty Carlo tree search terminologies (selection, rollout, exploration, and backpropagation) in a brief manner.

\begin{enumerate}
    \item State $S$: Each node or a state $S$ of the MCTS tree, provides a partial solution or a subset of possible node labeling for the concerned CO problem. Similar to the previous section, for the processing we maintain a binary vector $X_{v}$ where the $i^{th}$ entry is set to $1$ if $i^{th}$ node has already been labeled.
    \item Action, $a$ and Transition function, $\pi(S, a)$: An action $a$ means choosing a label (either $0$ or $1$) for the input graph node variable $x$. From each state $S$ multiple actions can be created by fixing the node labels of the unselected nodes from the point of view of $S$. Each action $a$ from the state $S$ also bears a transition probability $\pi(S, a)$ denoting the likelihood of taking action $a$ from $S$. $\pi(S,a)$ is approximated using a GNN.
    \item Reward, $r$: To calculate the reward for a state $S$, GNN is used. Using GNN, node probability distributions $P(\theta)$ are generated. Based on $S$, a subset of nodes have already been fixed, for the remaining unselected nodes' labels, $P(\theta)$ is used over a probability threshold $\beta$, e.g., for a node $v_{i}$, if $P_{i}(\theta) \geq \beta$, then $x_{i}=1$ else $x_{i}=0$. After approximating the labels of all the nodes, QUBO-formulated Hamiltonian is used to calculate the reward, $r = \sum_{i \leq j} x_{i}Q_{ij}x_{j}$.
\end{enumerate}

As it can be already understood, GNN plays a very important part of this design, now we state the mathematical formulation of GNN's forward pass along with the loss function that is used to update the parameters $\theta$.

  \begin{equation}
     \label{eqn:x_em}
         X_{em} = E(G)
    \end{equation}
    \begin{equation}
         \label{eqn:mu}
        \mu^{\prime} = GNN(G, X_{em})
    \end{equation}
    \begin{equation} \label{eqn:mu_final}
        \mu = f_{1}(X_{v}\theta_{1}^{T} + \mu^{\prime}\theta_{2})
    \end{equation}
    \begin{equation}
        \label{eqn:predict_mcts_gnn}
         P(\theta) = f_{2}(\mu\theta_{3}^{T})
    \end{equation}

The complete set of the mathematical formulation is presented from equation \ref{eqn:x_em} to equation \ref{eqn:predict_mcts_gnn}. First,
a set of node embedding vectors $X_{em}$ is generated (equation \ref{eqn:x_em}). Then, $X_{em}$ and input graph G is passed to GNN to generate
a set of node feature vectors $\mu^{\prime}$ (equation \ref{eqn:mu}). After that, more contextual information ($X_{v}$) is added with feature vector ($\mu^{\prime}$)
to calculate the complete node feature vectors $\mu$ (equation \ref{eqn:mu_final}). The final equation to generate the node probability distribution $P(\theta)$ is given
in \ref{eqn:predict_mcts_gnn}. Here the set of architecture parameters $\theta = \{\theta_{GNN} \in \mathbb{R}^{d_2 \times d_1}, \theta_{1} \in \mathbb{R}^{d_3 \times 1}, \theta_{2} \in \mathbb{R}^{d_3 \times d_2}, \theta_{3} \in \mathbb{R}^{ 1 \times d_3} \}$. We use $\theta_{GNN}$ to denote all the weight
parameters associated with the layers of GNN. $f_{1}$ is a ReLU activation function and $f_{2}$ is a sigmoid activation function to generate the probabilities. Similar to before
$(.)^{T}$ denotes the transpose from row vector to column vector or vice versa. $P(\theta)$ is also used to approximate $\pi(S, a)$. For a particular variable $x_{v}$ or input graph
node $v$, $P_{v}(\theta)$ will indiciate the likelihood of labeling $v$ to $1$ ($x_{v}=1, \pi(S, v=1)=P_{v}(\theta)$). Similarly, to label $x_{v}$ as $0$, we set $\pi(S,v=0)=1-P_{v}(\theta)$. From each state $S$, we create child nodes for the unselected variables for both of the labels.

To train the GNN architecture we use the formulation stated in equation \ref{eqn:loss_mcts_gnn} as the loss function. This function is quite similar to the equation presented
in \ref{eqn:pi_gnn_loss} except here we add manual perturbation by fixing the node labels to guide the searching. Here $X_{v,i}$ denotes the value of the $i^{th}$ entry in $X_{v}$. This
value is set to $1$ means, node $i$ has already been labeled and $0$ means it has not been labeled. Our underlying intuition behind this formulation is presented in the form of proposition
\ref{proposition:manual_perturb}.

 \begin{equation} \label{eqn:loss_mcts_gnn}
        L(\theta) =  \sum_{i \leq j, X_{V,i}=1, X_{v,j}=1} x_{i} Q_{ij}x_{j} + \sum_{i \leq j, X_{V,i}=1, X_{v,j}=0} x_{i} Q_{ij} p(\theta_{j}) + \sum_{i \leq j,X_{V,i}=0, X_{v,j}=0} p(\theta_{i}) Q_{ij} p(\theta_{j})
    \end{equation}

\begin{proposition}[Manual perturbation to avoid local minimas] \label{proposition:manual_perturb}
Through manual perturbation, we enforce the training to avoid various local minimas by tackling noises (different sets of node labels while updating the parameters) resulting in more robust architecture and improved
performance in terms of reducing constraint violations.
\end{proposition}

Now, we discuss the terminologies associated with MCTS through the following points,

\begin{enumerate}
	\item Selection and Exploration: In equation \ref{eqn:selection}, we present the greedy metric, Upper Confidence Bound (UCB) to measure the average reward obtainable for a child state $C_{i}$ with respect to its parent state $S$ combining its transition likelihood ($\pi$) of being selected. Here $C_{i}.w$ and $C_{i}.v$ denote the total amount of reward accumulated in state $C_{i}$ and the number of times $C_{i}$ has been visited respectively. $\alpha$ denotes a hyperparameter, $\pi(S, a)$ denotes the transition probability to state $C_{i}$ from $S$ reported by GNN. $S.v$ denotes the total number of times state $S$ was visited. $log$ denotes mathematical logarithmic operation. The node with the highest UCB value is selected to explore in its subtree. When a leaf node is reached in this manner, we start the rollout phase.

         \begin{equation} \label{eqn:selection}
            UCB(C_{i})= \frac{C_{i}.w}{C_{i}.v} + \alpha * \pi(S, a) * \sqrt{\frac{log(S.v)}{C_{i}.v}}
    \end{equation}

	\item Rollout: In the rollout phase, using GNN, we conduct the training for multiple epochs to approximate the node labels for the unlabeled nodes and calculate the rewards. We have already discussed how GNN is used to generate the node probability distributions in the previous paragraphs.
	\item Backpropagation: After the rollout phase we enter the backpropagation phase of MCTS. In this phase, we update the state variables, $v$ and $w$ for the path of the root to the current leaf state of the search tree. For all the nodes in the path we increment the visiting attribute by $1$ and add the reward to $w$ approximated by GNN.
\end{enumerate}

\section{Evaluation} \label{section:evaluation}
In this section, we empirically evaluate GRL and MCTS-GNN with simple GNN in addressing the Max-Cut problem for the graphs of different densities. The graphs were randomly generated and all the graphs were undirected. For a given graph of $n$ nodes, there can be at most $\frac{n \times (n-1)}{2}$ edges where any two nodes do not have multi-edges among them. Our prepared graph dataset can be found in \footnote{\url{https://github.com/rizveeredwan/learning-graph-structure/tree/main/Graph\%20Dataset}}. We conduct the experiments based on three metrics - the number of satisfied constraints, scalability and required time. All the experiments were conducted on a 64-bit machine having AMD Ryzen 9 5950x 16-Core Processor x 32, 128 GB RAM, and 24 GB NVIDIA GeForce RTX 3090 GPU. All the implementations were done in Python language based on the blocks provided by deep-learning libraries Pytorch\citep{paszke2019pytorch}, Pytorch Geometric \citep{fey2019fast} and LabML \citep{labml}.

\subsection{Architecture Description, Hyperparameter Setup}
In this section, we present the description of the architectures along with the value of the hyperparameters that are used during the training. The description is presented in the following manner,

\begin{enumerate}
    \item Architecture Description: To present the discussion we mostly use the variables stated in the respective sections.

    Our simple GNN architecture follows a similar definition stated in \cite{schuetz2022combinatorial}. It consists of two layers of Graph Convolutional Architecture (GCN) and a sigmoid layer to generate the node probabilities. The first GCN layer's node feature vectors are propagated through a non-linear ELU activation and a dropout layer before passing to the second layer of GCN. Let us assume the node feature vector size of the $1^{st}$ and $2^{nd}$ layers of GCN are $d_{1}$ and $d_{2}$. As mentioned in the base article, we follow a similar setup. If the number of nodes $n \geq 10^{5}$, then $d_{1}=\sqrt[3]{n}$, else $d_{1}=\sqrt{n}$. Also, $d_{2} = \frac{d_1}{2}$.

    To implement GRL we took inspiration from the study presented in \cite{drori2020learning}. Here, we had three layers of GAT as the encoding architecture and a sigmoid layer to generate initial node probabilities. The attention-based decoding formulation has been presented in Section \ref{section:grl}. To set the dimension of the node feature vectors, we follow a similar strategy stated in the previous paragraph based on the number of nodes of the input graph, $n$. So, if $n \geq 10^{5}$m then, $d_{1}=\sqrt[3]{n}$ else, $d_{1}=\sqrt{n}$ and followup $d_2 = \lceil \frac{d_1}{2} \rceil$. We have used a single attention-head mechanism.

    In MCTS-GNN, we train a single GNN in different rollout phases by applying different manual labeling perturbations. The setup is completely similar to the discussion already presented above. So, if $n \geq 10^{5}$, then  $d_{1}=\sqrt[3]{n}$ else, $d_{1}=\sqrt{n}$. $d_2 = \lceil \frac{d_1}{2} \rceil$ and $d_{3}=1$.
     \item Training Description: To train the architectures, we use Adam optimizer for each of the architectures. We use patience value to implement fuzzy early stopping for all of them. As RL-based setups inherently exhibit abrupt behavior (frequent ups and downs in terms of reward variation) we keep a comparatively higher patience value for GRL and MCTS-GNN compared to PI-GNN. As per our setups, we eventually observed almost linear variation in terms of objective functions' values at some phase of the training epochs for all the experimented architectures for different graph inputs. This supports the validity of the chosen hyperparameters for early stopping. Another reason, behind setting a higher patience for RL-based setups is that, here we set the bar over the actual integer reward value compared to the fractional loss variation in PI-GNN. Inherently, the integer rewards should fluctuate more compared to the fractional loss values.

     For PI-GNN, we simply choose learning rate $lr=10^{-4}$ with a patience value of $\tau=100$ denoting if there is no loss objective improvement compared to the current best value observed for a consecutive $100$ epochs, we stop the training. As stated previously, we apply fuzzy early stopping here. For GRL, we choose a learning rate of $0.001$ ($lr$) for both encoder-decoder and a patience value of $700$ ($\tau$). We also experimented with a lesser learning rate ($0.0001$) for GRL but could not improve the performance in terms of satisfying the number of constraints. For, MCTS-GNN, we also chose the learning rate $lr$ to $0.001$ for the GNN architecture with the patience value of $100$. But here we also had another early stopping criterion denoting the improvement in the reward objective. Here we set another patience value $\tau^{\prime}$ to $700$ denoting if there is no reward improvement compared to the best reward observed till the current iteration, the MCTS-GNN algorithm terminates.
\end{enumerate}

\subsection{Comparison in terms of number of satisfied constraints}
In this section, we present the number of satisfied constraints observed in each architecture, PI-GNN, GRL and MCTS-GNN for the graphs of different intensities. The complete result is shown in Table \ref{tab:reward}. In this study, this metric is denoted as the reward in terms of RL-based formulation. For GRL and MCTS-GNN we present the best reward observed throughout the complete training. For PI-GNN, we record the least loss and the corresponding probability distribution to approximate the node labels based on the threshold $\beta=0.5$ - the node's probability exceeding $\beta$ gets labeled as $1$. These approximated node labels are used to calculate the QUBO-formulated Hamiltonian function. In the last two columns of Table \ref{tab:reward}, we present how GRL and MCTS-GNN's performance has improved compared to PI-GNN in percentage. Positive value means, the result has improved and the negative means the opposite.

\begin{table}[!ht]
    \centering
    \begin{tabular}{c|c|c|c|c|c|c}
    \hline
    \# Nodes & \# Edges & PI-GNN & GRL & MCTS-GNN & GRL & MCTS-GNN\\
    & & & & &  vs PI-GNN(\%)  &  vs PI-GNN (\%) \\
    \hline
    50 & 89 & 72 & 75 & 76* & 4.2 & 5.56\\
    50 & 139 & 95 & 105 & 107 * & 10.53 & 12.63 \\
   50 & 499 & 276 & 301 & 314* & 10.14 & 13.78\\
   \hline
100 & 199 & 147 & 155 & 167* & 5.44 & 13.61\\
100 & 799 & 373 & 534 & 537* & 43.2 & 44\\
\hline
300 & 399 & 342 & 360 & 365* & 5.26 & 6.73\\
300 & 899 & 613 & 690 & 699* & 12.56 & 14.03\\
300 & 1299 & 747 & 932 & 947* & 24.77 & 26.77\\
\hline
500 & 799 & 622 & 672 & 694* & 8.04 & 11.58\\
500 & 1499 & 1007 & 1143 & 1158* & 13.51 & 15.0\\
500 & 5499 & 2689 & 3414 & 3535* & 26.96 & 31.46\\
\hline
700 & 1199 & 938 & 979 & 1023* & 4.37 & 9.06\\
700 & 1699 & 1288 & 1308 & 1360* & 1.55 & 5.59\\
700 & 4699 & 2422 & 2992 & 3202* & 23.53 & 32.2\\
\hline
1000 & 1299 & 1104 & 1112 & 1141* &  0.73 & 3.35\\
1000 & 3299 & 2204 & 2449 & 2525* & 6.9 & 14.56\\
1000 & 5299 & 3098 & 3716 & 3750* & 19.95 & 21.05\\
\hline
3000 & 3499 & 2956 & 3218* & 2996 & 8.86 & 1.35\\
3000 & 4499 & 3627 & 3907* & 3885 & 7.72 & 7.11\\
3000 & 6999 & 4841 & 5550 & 5622* & 14.65 & 16.13\\
    \hline

    \end{tabular}
    \caption{Number of Satisfied Constraints (Reward) for the graphs of different intensities for the Max-Cut problem. (*) denotes the best value observed.}
    \label{tab:reward}
\end{table}

From the presented result in Table \ref{tab:reward}, we can see that, the performance has comparatively improved than the base PI-GNN. In simple terms, upon applying RL-based formulation we improved the quality of the resultant output. If we investigate in a more detailed manner we can see that, for a particular graph having $n$ nodes, the performance generally improves more with it being denser or increased number of edges.

\subsection{Training Stability}
\begin{figure}
    \centering
    \begin{minipage}{0.32\textwidth}
        \includegraphics[width=\textwidth]{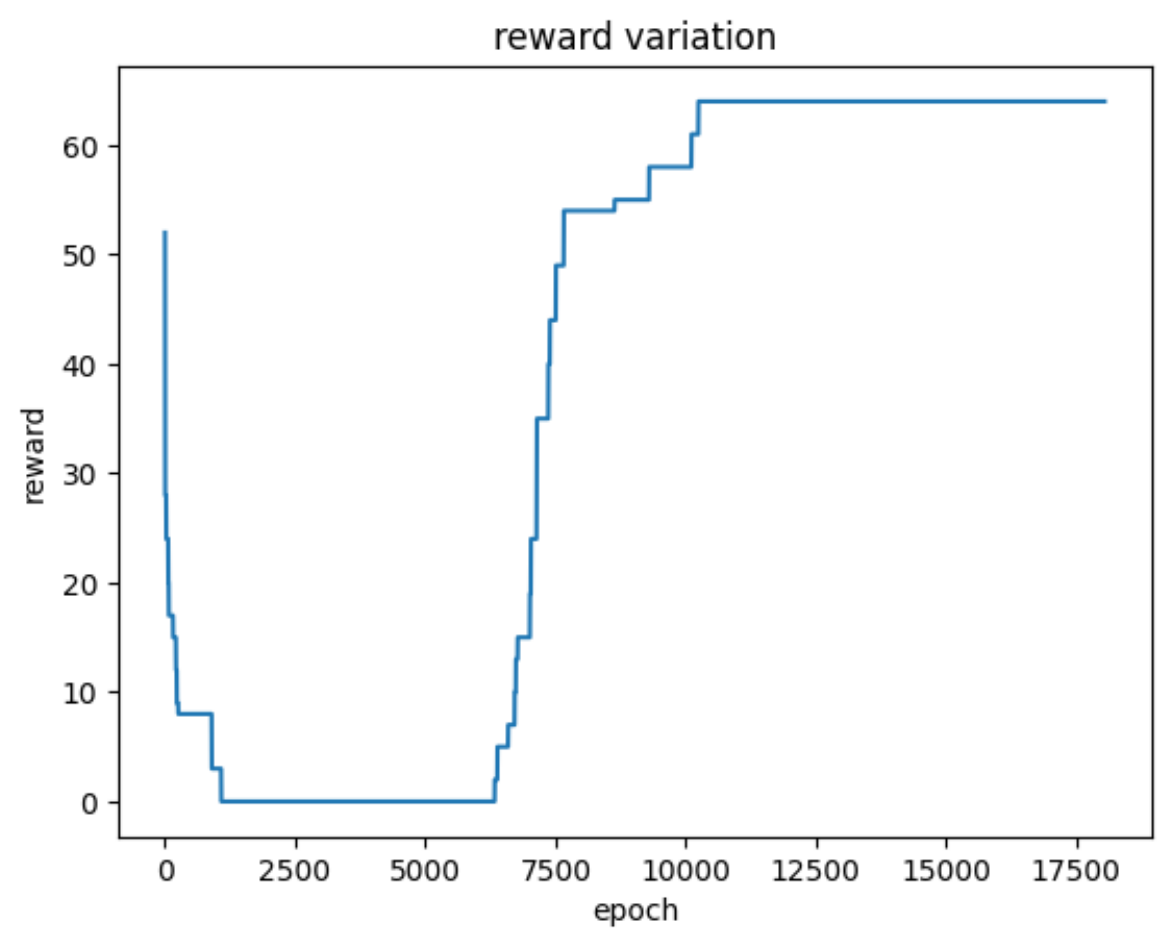}
        \caption{Reward variation curve for PI-GNN for the graph (50, 89)}
    \end{minipage}
    \begin{minipage}{0.32\textwidth}
        \includegraphics[width=\textwidth]{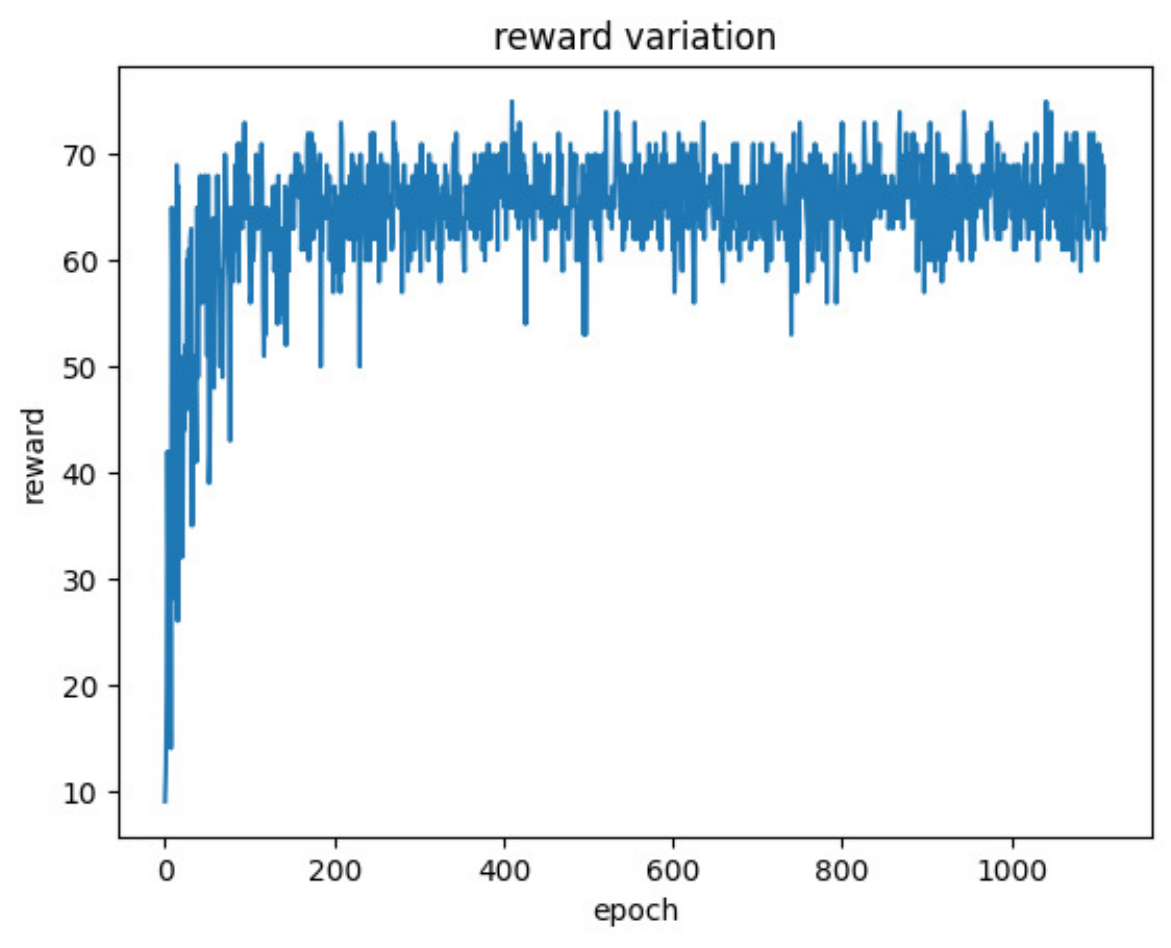}
        \caption{Reward variation curve for GRL for the graph (50, 89)}
    \end{minipage}
    \begin{minipage}{0.32\textwidth}
        \includegraphics[width=\textwidth]{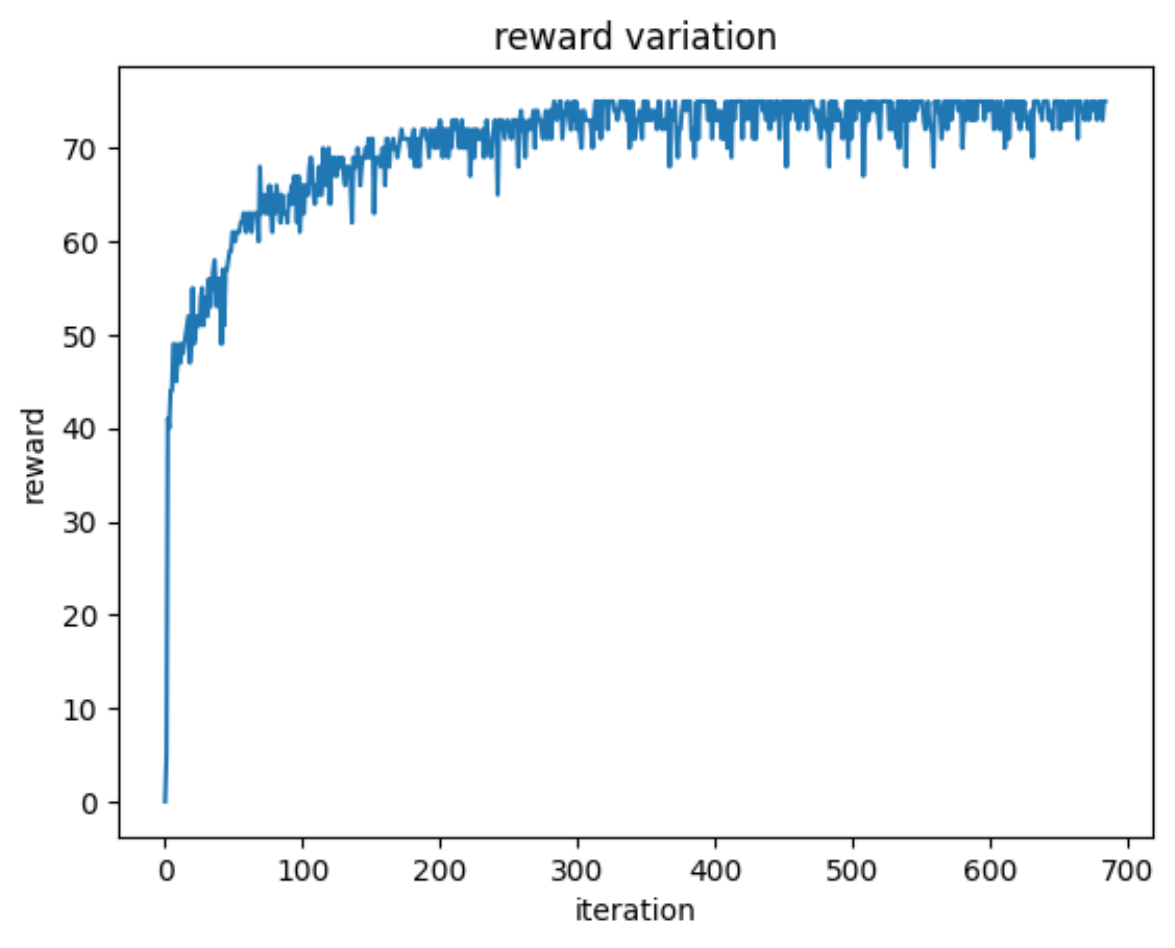}
        \caption{Reward variation curve for MCTS-GNN for the graph (50, 89)}
    \end{minipage}
    \label{fig:50_89}
\end{figure}

\begin{figure}
    \centering
    \begin{minipage}{0.32\textwidth}
        \includegraphics[width=\textwidth]{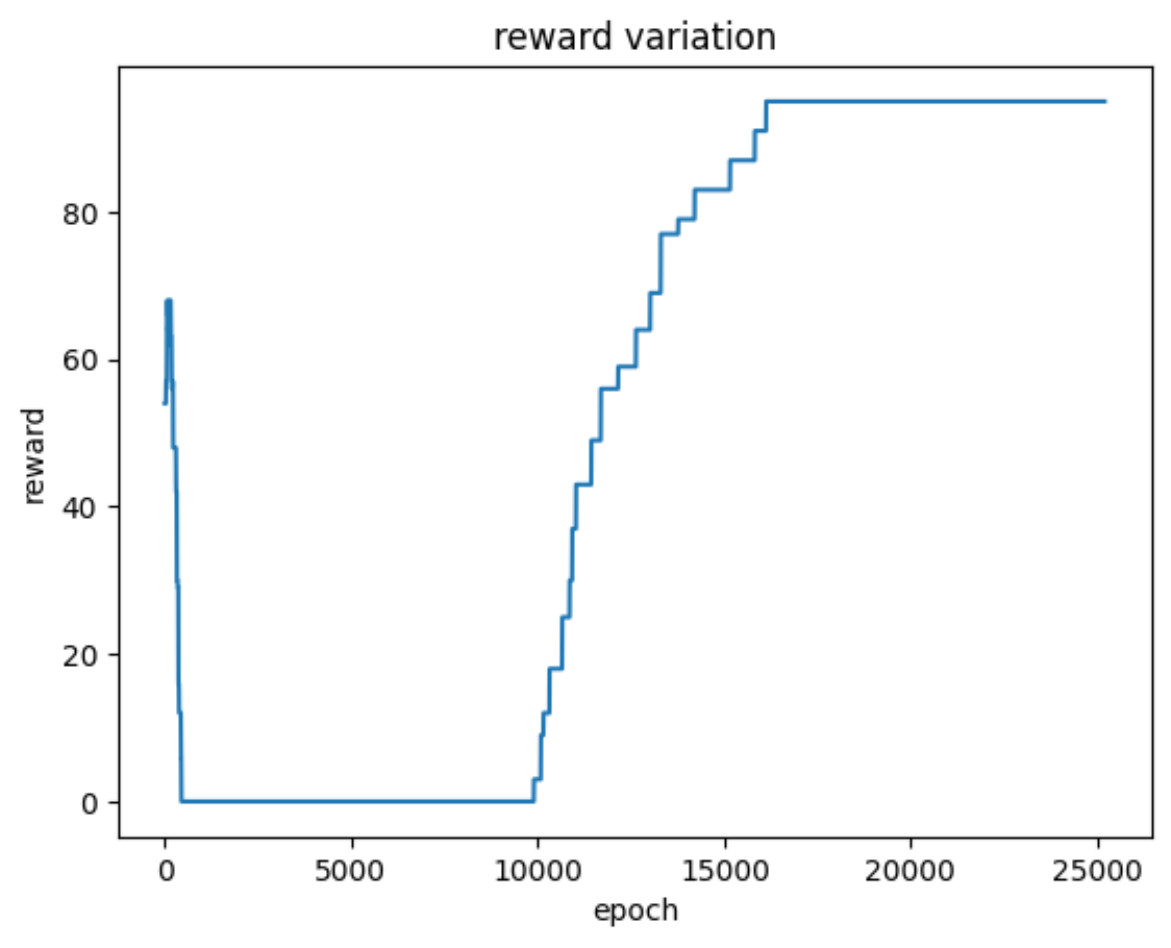}
        \caption{Reward variation curve for PI-GNN for the graph (50, 139)}
    \end{minipage}
    \begin{minipage}{0.32\textwidth}
        \includegraphics[width=\textwidth]{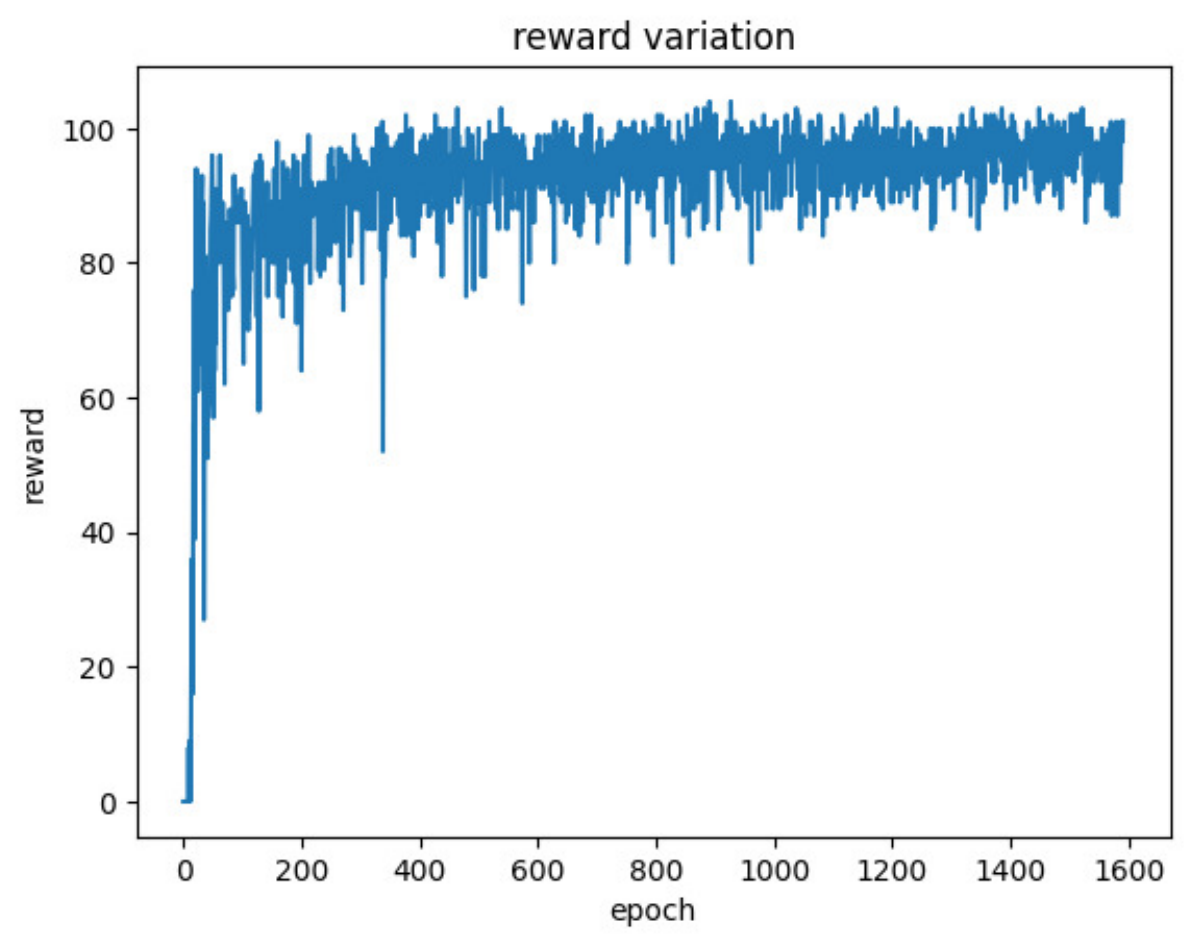}
        \caption{Reward variation curve for GRL for the graph (50, 139)}
    \end{minipage}
    \begin{minipage}{0.32\textwidth}
        \includegraphics[width=\textwidth]{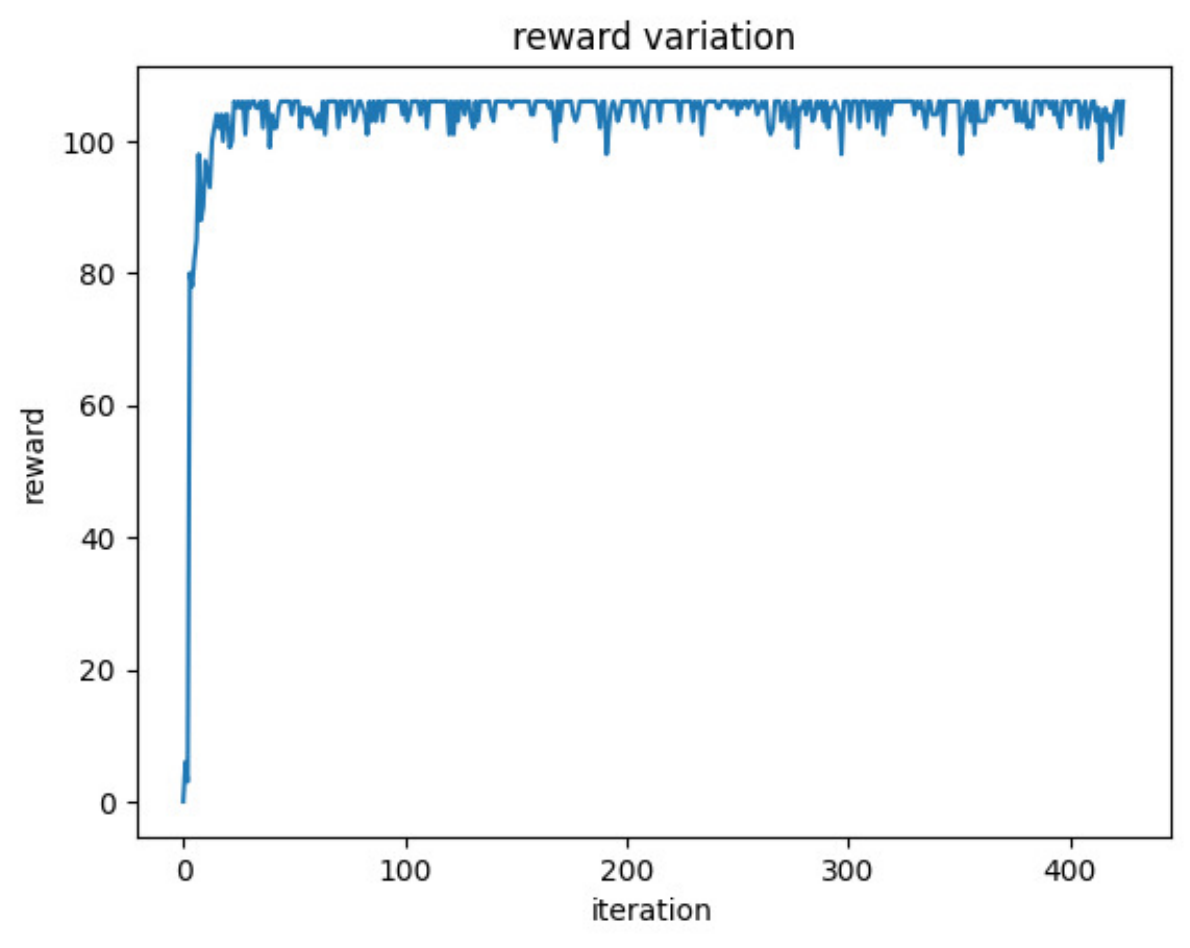}
        \caption{Reward variation curve for MCTS-GNN for the graph (50, 139)}
    \end{minipage}
    \label{fig:50_139}
\end{figure}

\begin{figure}
    \centering
    \begin{minipage}{0.32\textwidth}
        \includegraphics[width=\textwidth]{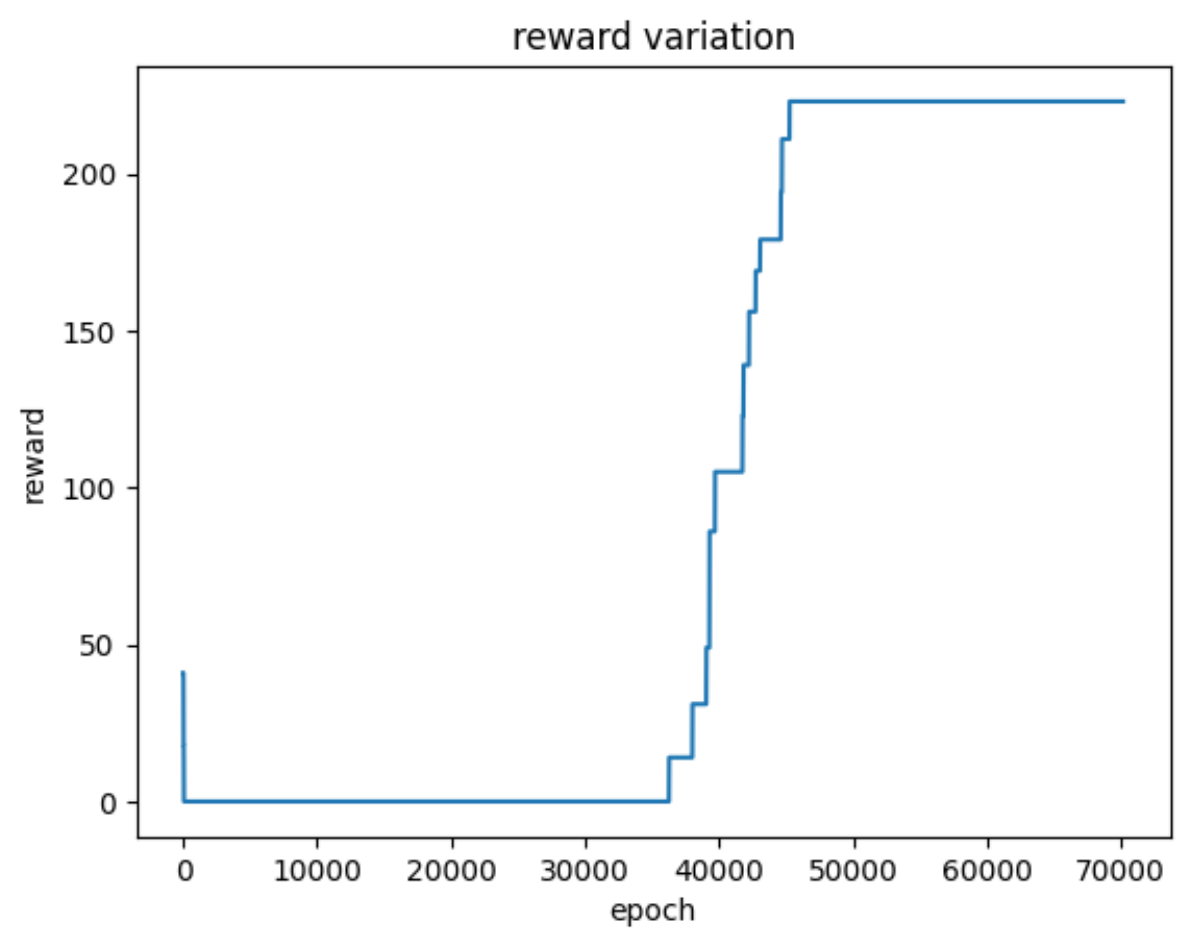}
        \caption{Reward variation curve for PI-GNN for the graph (50, 499)}
    \end{minipage}
    \begin{minipage}{0.32\textwidth}
        \includegraphics[width=\textwidth]{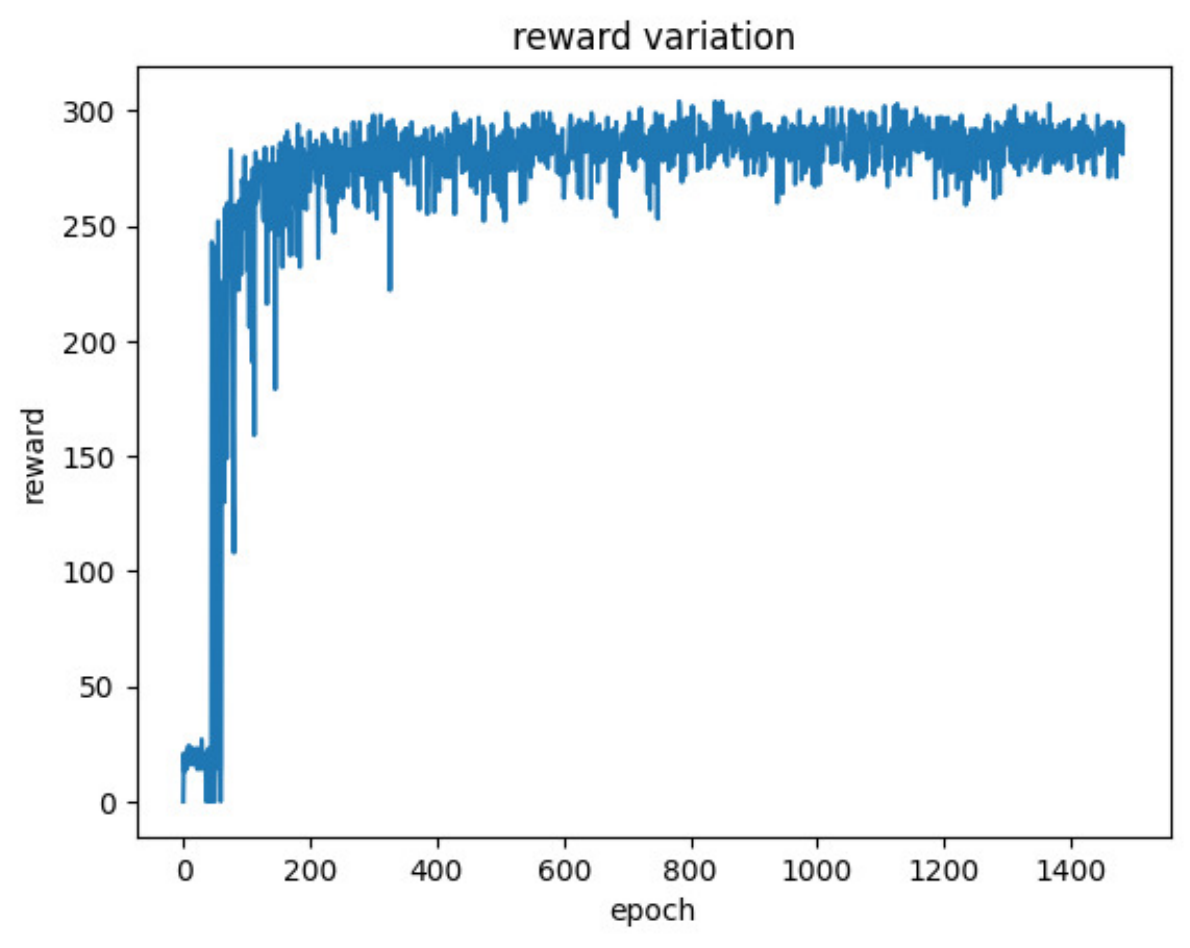}
        \caption{Reward variation curve for GRL for the graph (50, 499)}
    \end{minipage}
    \begin{minipage}{0.32\textwidth}
        \includegraphics[width=\textwidth]{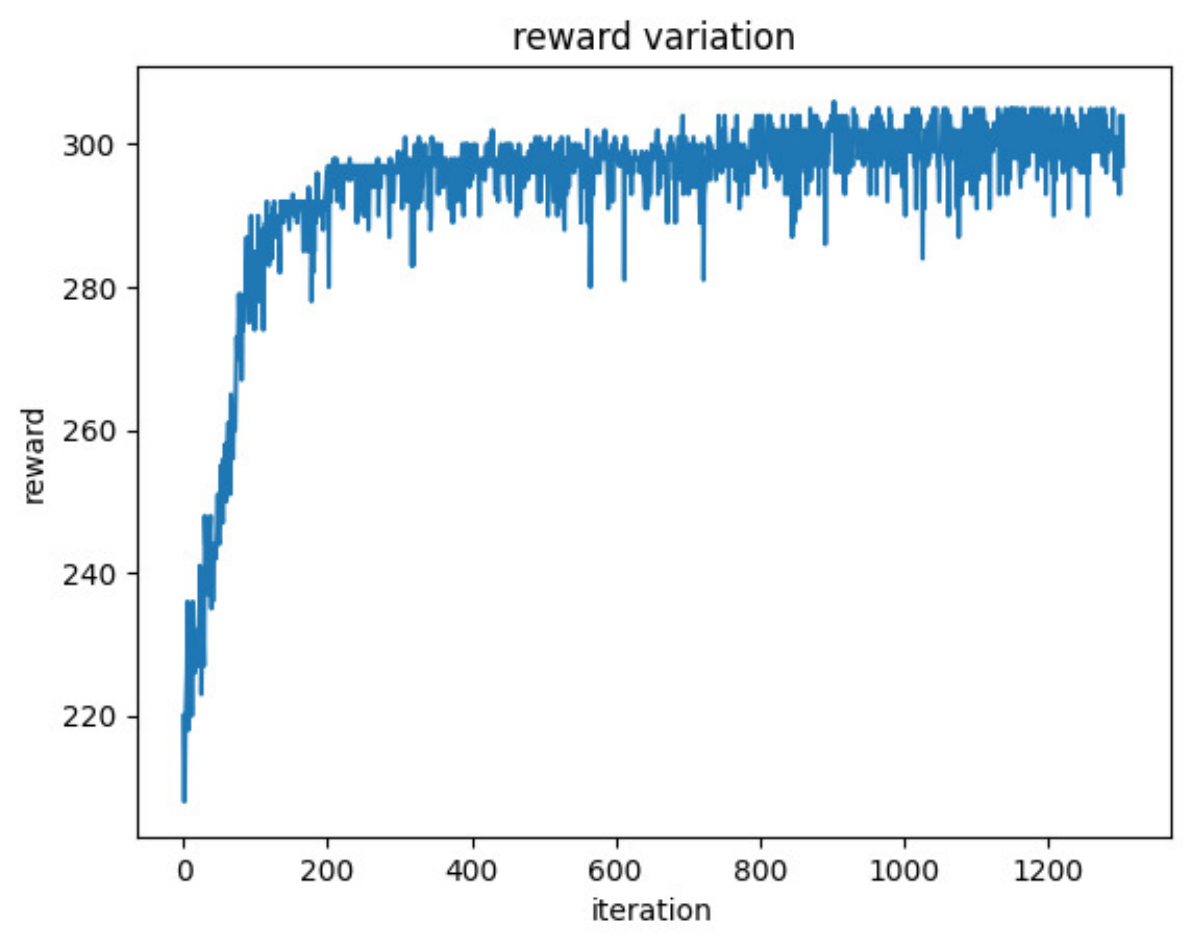}
        \caption{Reward variation curve for MCTS-GNN for the graph (50, 499)}
    \end{minipage}
    \label{fig:50_499}
\end{figure}

\begin{figure}
    \centering
    \begin{minipage}{0.32\textwidth}
        \includegraphics[width=\textwidth]{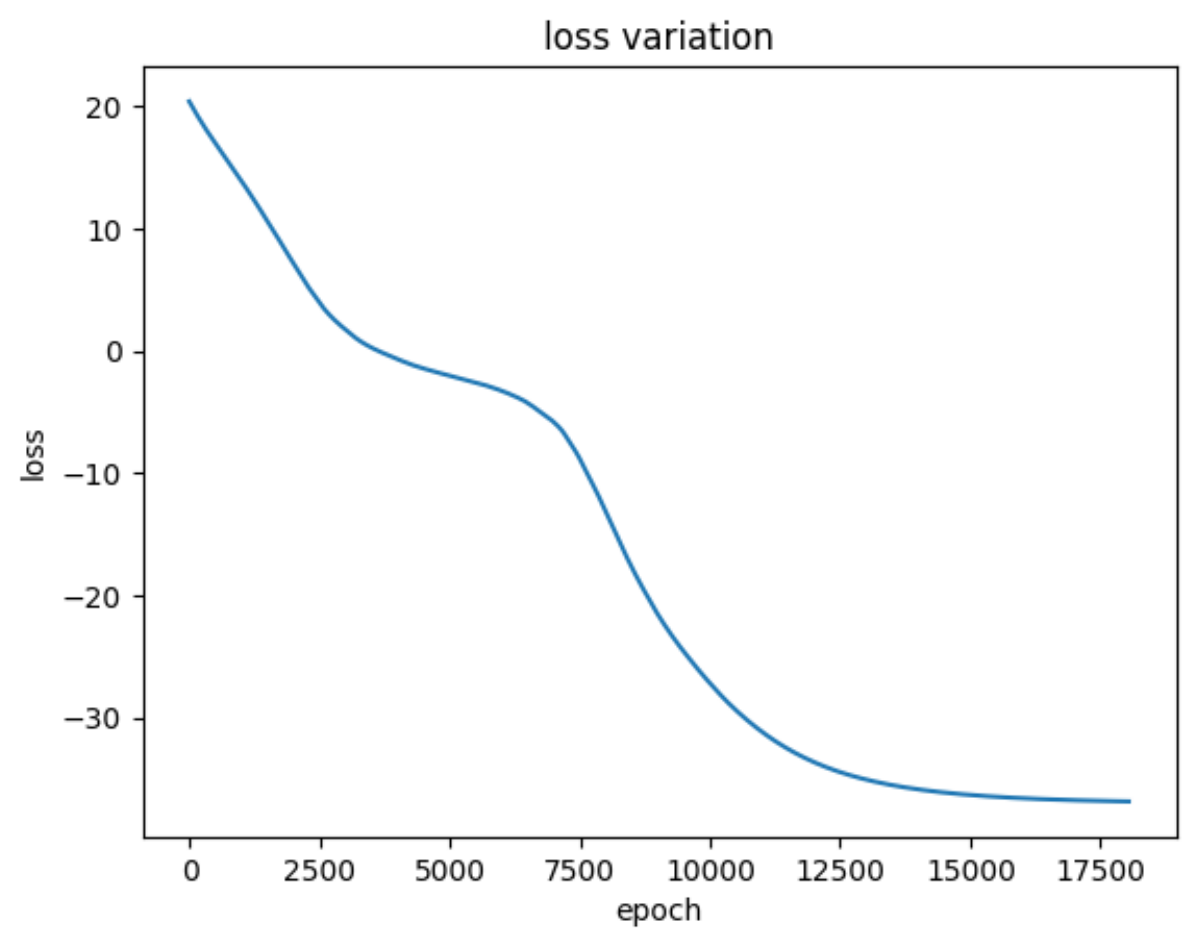}
    \caption{Loss variation curve for PI-GNN for the graph (50, 89)}
    \end{minipage}
     \begin{minipage}{0.32\textwidth}
        \includegraphics[width=\textwidth]{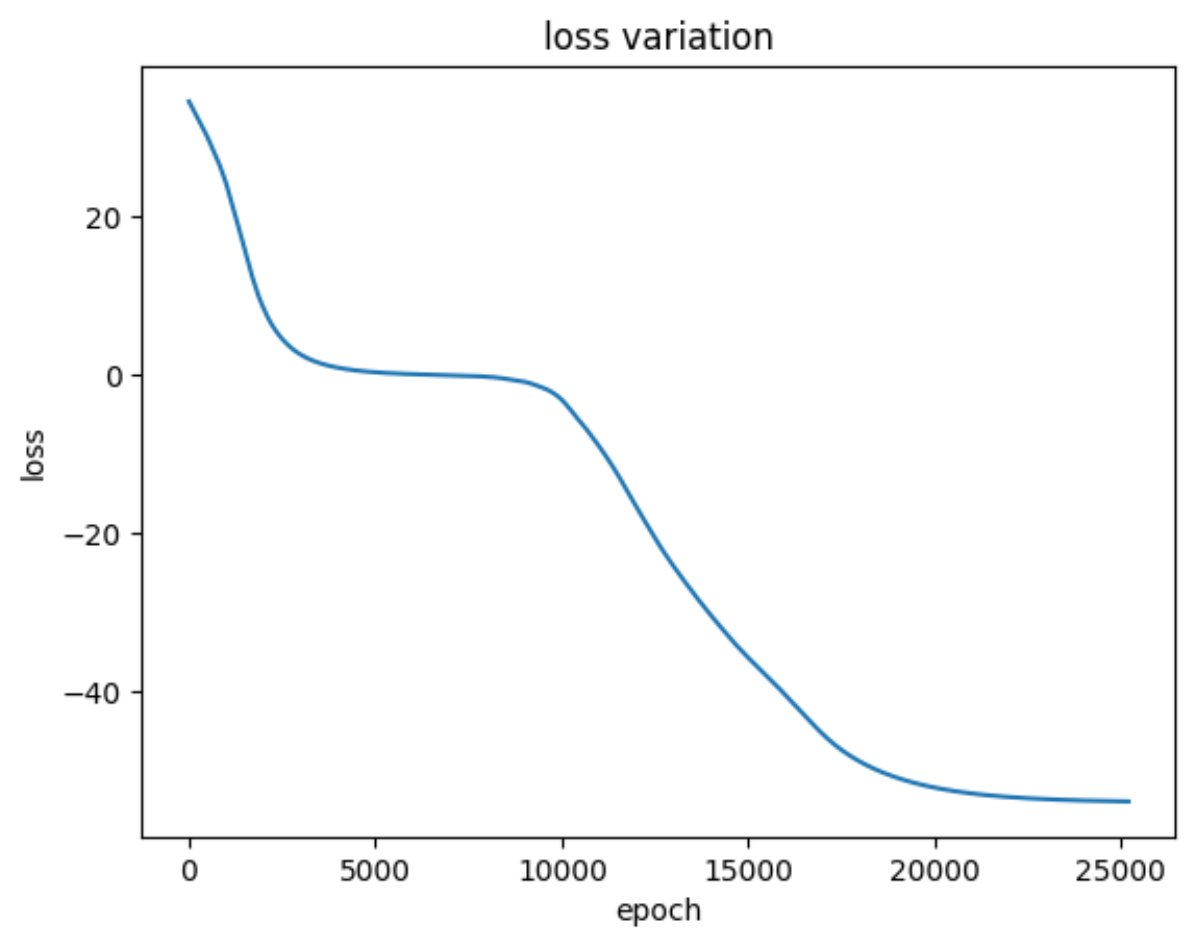}
    \caption{Loss variation curve for PI-GNN for the graph (50, 139)}
    \end{minipage}
     \begin{minipage}{0.32\textwidth}
        \includegraphics[width=\textwidth]{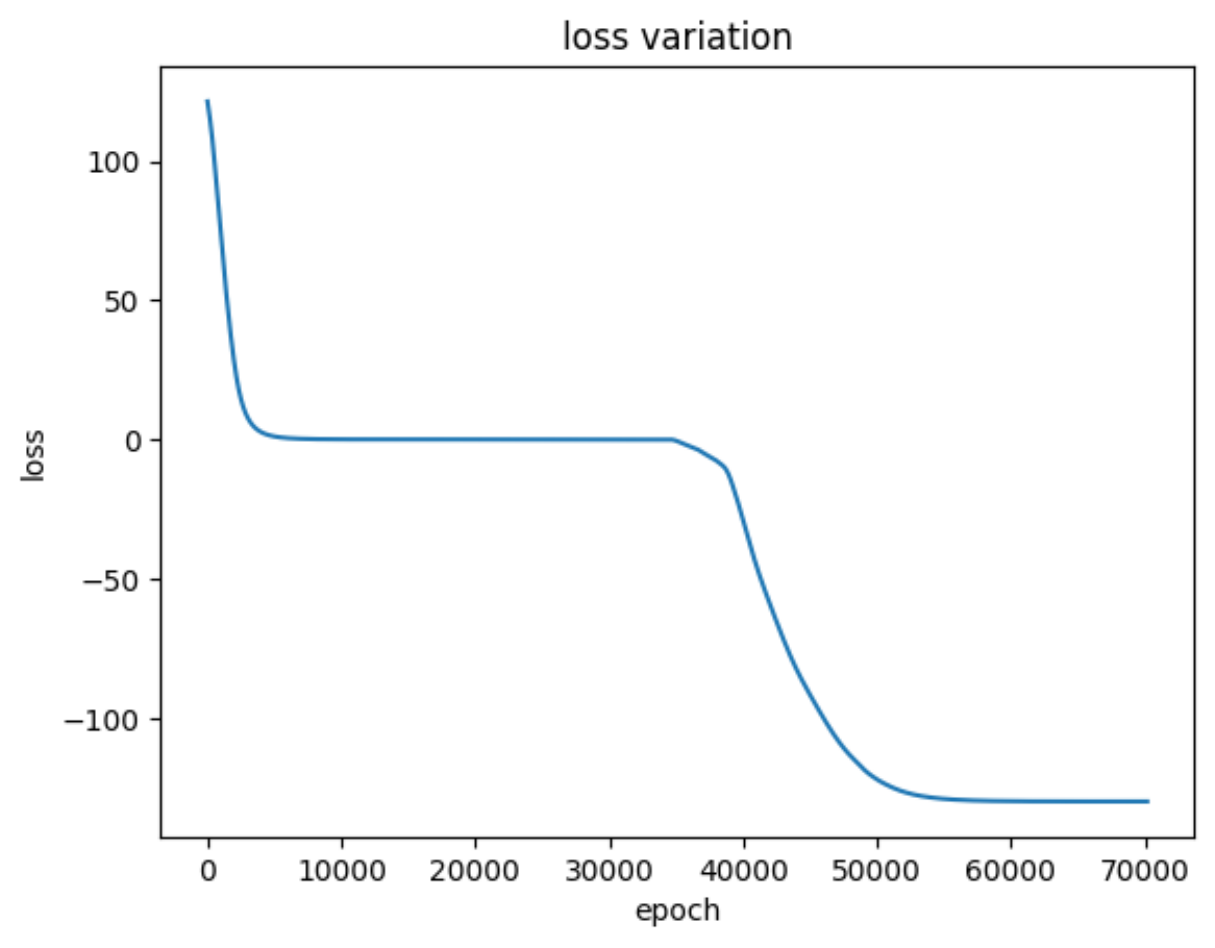}
    \caption{Loss variation curve for PI-GNN for the graph (50, 499)}
    \end{minipage}
    \label{fig:loss_pi_gnn}
\end{figure}

In this section, we mainly highlight the convergence status or training stability of the experimented architectures based on our set hyperparameters. We centralize our discussion based on three graphs of $50$ nodes with $89$, $139$, and $499$ edges respectively for all the architectures to understand the behavior transition from sparser to denser graphs. In Fig. \ref{fig:50_89} we present the reward variations for PI-GNN, GRL, and MCTS-GNN respectively for a graph of 50 nodes with 89 edges. In a similar manner, we present the reward variations for a graph of 50 nodes with 139 edges and a graph of 50 nodes with 499 edges for all the architectures in Fig. \ref{fig:50_139} and \ref{fig:50_499} respectively. To specifically understand the loss convergence status of PI-GNN we present Fig. \ref{fig:loss_pi_gnn} for the three 50 node graphs with 89, 139, and 499 edges respectively. Here, in the captions (50, 89) means a graph having 50 nodes and 89 edges.

The main observation that can be drawn from each is that the convergence behavior or almost linear variation is quite visible where our training has stopped for all the architectures. This supports the stability quality of the set hyperparameters, especially the patience values. Also, we do not provide all the other graphs' training convergence chart in this study, because they exhibit a similar pattern. All of our experiments are reproducible and can be regenerated by importing our official code repository.

\subsection{Complexity Analysis}
In this section, we concisely present the time complexity of each of the experimented architectures or more specifically discuss the computation-intensive segments for each.

PI-GNN is the most scalable and runtime efficient compared to GRL and MCTS-GNN due to the novelties of GNN architecture. In simpler terms, to process a graph having $n$ nodes to generate a $d$ dimensional node feature vectors through $L$ layers of GNN, the big-O complexity is $\mathcal{O}(Lnd^{2})$\footnote{Here our assumption is that, the graph is sparse so, the number of edges $E=\mathcal{O}(n)$}. GRL also uses a GNN-variation (GAT) to encode the input graph. After that, it applies greedy selection at each iteration to pick a node with the highest attention value reported by the decoder. The additional complexity of this selection portion is $\mathcal{O}(nLogn)$ - maintaining a max heap structure, per selection takes $O(logn)$ complexity. MCTS-GNN also uses a single GCN architecture in the rollout phases to build the search tree. So, a similar complexity of GNN is automatically added to the overall complexity. It also takes multiple attempts to train the same GNN again by varying inputs through manual perturbation of labelings. So, this incurs additional costs also. But our experiments suggest that only a very few times, GNN is trained exhaustively, and the other times, it generally runs for a very small number of epochs (less than $1000$ in our experiments) and reaches the early stopping criterion. So, in summary, PI-GNN is the most scalable. The complexity of GRL increases with the graphs becoming larger and the complexity of MCTS-GNN increases with the number of iterations to run to expand the search tree.

Apart from these theoretical aspects, other constraints, e.g., step size or learning rate are also quite important and play a crucial role in converging the objective loss functions. If it takes a good time for the convergence, then the overall training time increases also. Based on our observations, generally, we found MCTS-GNN to take the most amount of time due to the expansion of the search tree. GRL takes way less time than MCTS-GNN and provides a competitive performance against PI-GNN - where PI-GNN takes a longer number of epochs to be trained, GRL takes way less number of epochs where there lies a significant amount of processing in each epoch by selecting the nodes in a greedy manner. PI-GNN sometimes gives worse performance than GRL, especially in denser graphs where it might need a good amount of time to be converged. In Table \ref{tab:time}, we present some results, each value is denoted in seconds. But, a point to be mentioned is that, based on loss convergence status, in multiple trials, the runtime can vary significantly.

\begin{table}[]
    \centering
    \begin{tabular}{c|c|c|c|c}
       Node  & Edge & PI-GNN & GRL & MCTS-GNN\\
       \hline
       50 & 89 & 698 & 211 & 1066 \\
       50 & 139 & 691 & 184 & 835 \\
       50 & 499 & 3074 & 439 & 1834\\
       \hline
       100 & 199 & 317 & 347 & 5147\\
       100 & 799  & 2668 & 602 & 8479\\
    \hline
    300 & 399  & 398 & 427 & 1027\\
    300 & 899  & 667 & 862 & 5789\\
    300 & 1299  & 1837 & 1789 & 9035\\
    \hline
    500 & 799  & 478 & 679 & 1507\\
    500 & 1499  & 1478 & 1025 & 6478 \\
    500 & 5499  & 2478 & 2247 & 8798 \\
    \hline
    700 & 1199  &  932 & 725 & 2017 \\
    700 & 1699  &  1027 & 879 & 4789 \\
    700 & 4699  & 2104 & 2578 & 9786 \\
    \hline
    1000 & 1299  & 1265 & 1681 & 10510\\
    1000 & 3299  & 2333 & 1414 & 15146\\
    1000 & 5299  & 2017 & 2768 & 22147\\
    \hline
    3000 & 3499  & 1879 & 2147 & 8978 \\
    3000 & 4499  & 2998 & 2378 & 13147 \\
    3000 & 6999  & 5014 & 4789 & 24789 \\
    \hline
    \end{tabular}
    \caption{Training time in seconds for PI-GNN, GRL, and MCTS-GNN for different graphs}
    \label{tab:time}
\end{table}

\section{Discussion and Conclusion} \label{section:conclusion}
In this study, we extend the work of \cite{schuetz2022combinatorial}. Our main contributions are - identifying a crucial issue during training of PI-GNN while early stopping and experimenting with RL-based formulations to understand the performance in terms of the number of satisfied constraints. In GRL, we experiment with a QUBO-formulated Hamiltonian as a generic reward function, and in MCTS-GNN we apply the Monty Carlo tree search with GNN guided by manual perturbation of node labeling. Based on our experiments, we found that RL-based setups generally give comparatively better performance in terms of satisfied constraints than the simple PI-GNN setup with additional incurring runtime or processing costs. So, our summarized observation over \cite{boettcher2023inability} and \cite{angelini2023modern} is that PI-GNN \citep{schuetz2022combinatorial} is quite scalable and provides a moderate performance in terms of the number of satisfied constraints which can be improved by enforcing RL-based formulations. In the next phase of our work, we plan to investigate more RL-based sophisticated formulations in terms of addressing CO problems considering scalability, training criteria, graph representation, etc.

\bibliographystyle{unsrtnat}


\end{document}